\documentstyle[eaclap]{article}

\title{\vspace{-0.5in}ProFIT: Prolog with Features, Inheritance and Templates}
\author{Gregor Erbach\\
Universit\"at des Saarlandes\\
Computerlinguistik\\
D-66041 Saarbr\"ucken, Germany\\
e-mail: {\small \tt erbach@coli.uni-sb.de}\\
{\sc url:} {\small \tt http://coli.uni-sb.de/\~{}erbach/}\\
CMP-LG e-print archive: cmp-lg/9502003
}

\begin{document}

\maketitle
\vspace{-0.5in}
\begin{abstract}
ProFIT is an extension of Standard Prolog
with Features, Inheritance and
Templates. ProFIT allows the programmer or grammar developer to declare
an inheritance
hierarchy, features and templates. Sorted feature terms can be used
in ProFIT programs
together with Prolog terms to provide a clearer description
language for linguistic structures.
ProFIT compiles all sorted feature terms
into a Prolog term representation, so that the built-in Prolog term
unification can be used for the unification of sorted feature structures,
and no special unification algorithm is needed.
ProFIT programs are compiled into Prolog programs, so that no meta-interpreter
is needed for their execution. ProFIT thus provides
a direct step from grammars developed with sorted feature terms to
Prolog programs usable for practical NLP systems.
\end{abstract}

\section{Introduction}

There are two key ingredients for building an NLP system:
\begin{itemize}
\item a linguistic description
\item a processing model (parser, generator etc.)
\end{itemize}

In the past decade, there have been diverging trends in the area of
linguistic descriptions and in the area of processing models. Most
large-scale linguistic descriptions make use of sorted feature formalisms,%
\footnote{Sorted feature structures are sometimes referred to as {\em typed}
          feature structures, e.g. in Carpenter's ``Logic of Typed
          Feature Structures.'' We follow the usage in Logic Programming
          and the recent {\sc hpsg} literature.}
but implementations of these formalisms are in general too slow for
building practically usable NLP systems.
Most of the progress in constructing efficient parsers and generators
has been based on logic grammars that make use of ordinary Prolog terms.
We provide a general tool that brings together these developments by
compiling sorted feature terms into a Prolog term representation, so that
techniques from logic programming and logic grammars can be used
to provide efficient processing models for sorted feature grammars.

In this introductory section, we discuss the advantages of sorted
feature formalisms, and of the logic grammar paradigm, and show how
the two developments can be combined. The following sections describe
the ProFIT language which provides sorted feature terms for Prolog,
and its implementation.

\subsection{Grammar Development in Sorted Feature Formalisms}

Sorted feature formalisms are often used for the development of large-coverage
grammars, because they are very well suited for a structured description
of complex linguistic data. Sorted feature terms have several advantages over
Prolog terms as a representation langauge.

\begin{enumerate}
\item They provide a compact notation. Features that are not instantiated
      can be omitted; there is no need for anonymous variables.
\item Features names are mnemonic, argument positions are not.
\item Adding a new feature to a sort requires one change in a declaration,
      whereas adding an argument
      to a Prolog functor requires changes (mostly insertion of anonymous
      variables) to every occurence of the functor.
\item Specification of the subsort relationship is more convenient than
      constructing Prolog terms which mirror these subsumption relationships.
\end{enumerate}

Implementations of sorted feature formalisms such as
TDL \cite{Krieger:Schaefer:94},
ALE \cite{Carpenter:93},
CUF \cite{Doerre:Dorna:93},
TFS \cite{Emele:Zajac:90}
and others have been used successfully for the development and testing
of large grammars and lexicons, but they may be too slow for actual use in
applications because they are generally built on top of Prolog or LISP,
and can therefore not be as efficient as the built-in unification of
Prolog. There are a few logic programming langauges,
such as LIFE \cite{Kaci:Lincoln:89} or Oz \cite{Smo:Hen:Wue:95}, that
provide sorted feature terms, but no commercial implementations of these
languages with efficient compilers are yet available.

\subsection{Efficient Processing based on Logic Grammars}

Much work on efficient processing algorithms has been done in the logic
grammar framework. This includes work on

\begin{itemize}
\item Compiling grammars into efficient parsers and generators:
      compilation of DCGs into (top-down) Prolog programs,
      left-corner parsers (BUP), LR parsers, head-corner parsers,
      and semantic-head driven generators.
\item Use of meta-programming for self-monitoring to ensure generation
      of unambiguous utterances \cite{Neumann:Noord:92}
\item Work in the area of Explanation-Based Learning (EBL) to learn
      frequently used structures \cite{Samuelsson:94}
\item Tabulation techniques, from the use of well-formed substring tables
      to the latest developments in Earley deduction, and memoing techniques
      for logic programming \cite{Neummy:94}
\item Work based on Constraint Logic Programming (CLP) to provide processing
      models for principle-based grammars \cite{Matiasek:94}
\item Using coroutining (dif, freeze etc.) to provide more efficient
      processing models
\item Partial deduction techniques to produce more efficient grammars
\item Using Prolog and its indexing facilities to build up a lexicon database
\end{itemize}

Since much of this work involves compilation of grammars into Prolog programs,
such programs can immediately benefit from any improvements in Prolog compilers
(for example the tabulation provided by XSB Prolog can provide a more efficient
implementation of charts) which makes the grammars more usable for NLP
systems.

\subsection{Combining Logic Grammars and Sorted Feature Formalisms}

It has been noted that first-order Prolog terms provide the equivalent
expressive power as sorted feature terms \cite{Mellish:92}. For example,
Carpenter's typed feature structures \cite{Carpenter:92} can easily be
represented as Prolog terms, if the restriction is given up that the
sort hierarchy be a bounded complete partial order.

Such compilation of sorted feature terms into Prolog terms has been
successfully used in the Core Language Engine (CLE) \cite{Alshawi:91}
and in the
Advanced Linguistic Engineering Platform (ALEP), \cite{ET61}.%
\footnote{Similar, but less efficient compilation schemes are used
          in Hirsh's P-PATR \cite{Hirsh:86}
          and Covington's GULP system \cite{Covington:89}.}
ProFIT extends
the compilation techniques of these systems through the handling of
multi-dimensional inheritance \cite{Erbach:94b}, and makes them generally
available for a wide
range of applications by translating programs (or grammars) with sorted
feature terms into Prolog programs.

ProFIT is not a grammar formalism, but rather extends any grammar
formalism in the logic grammar tradition with the expressive power
of sorted feature terms.

\section{The ProFIT Language}

The set of ProFIT programs is a superset of Prolog programs.
While a Prolog program consists only of definite clauses (Prolog is
an untyped language), a ProFIT program
consists of datatype declarations and definite clauses. The clauses of a ProFIT
program
can make use of the datatypes (sorts, features, templates and finite domains)
that are introduced in the declarations. A ProFIT
program consists of:

\begin{itemize}
\item Declarations for sorts
\item Declarations for features
\item Declarations for templates
\item Declarations for finite domains
\item Definite clauses
\end{itemize}

\subsection{Sort Declarations} \label{sort-declarations}

In addition to unsorted Prolog terms, ProFIT allows sorted feature terms,
for which the sorts and features must be declared in advance.

The most general sort is {\tt top}, and all other sorts must be subsorts
of {\tt top}. Subsort declarations have the syntax given in
(\ref{subsort-decl}).
The declaration states that all $Sub_i$ are subsorts of $Super$, and
that all $Sub_i$ are mutually exclusive.

\begin{equation}
Super > [Sub_1,\ldots ,Sub_n].  \label{subsort-decl}
\end{equation}

It is also possible to provide subsorts that are not mutually exclusive, as
in (\ref{multi1}), where one subsort may be chosen from each of the
``dimensions'' connected by the $*$ operator \cite{Erbach:94b}.

\begin{equation}
\begin{array}{lll}
Super & > & [Sub_{1.1},\ldots ,Sub_{1.n}]\ \ * \\
      &   & \hspace*{15mm} \vdots  \\
      &   &  [Sub_{k.1},\ldots ,Sub_{k.m}]
\end{array}
\label{multi1}
\end{equation}

Every sort must only be defined once, i.e. it can appear only once
on the left-hand side of the connective $>$.

The sort hierarchy must not contain any cycles, i.e. there must be
no sorts $A$ and $B$, such that $A \neq B$, and $A > B > A$.

The immediate subsorts of {\tt top} can be declared to be extensional.
Two terms which are of an extensional sort
are only identical if they have a most specific sort (which has
no subsort), and if all features are instantiated to ground terms. If
a sort is not declared as extensional, it is intensional. Two intensional
terms are identical only if they have been unified.

\subsection{Feature Declarations}

Unlike unsorted feature formalisms (such as {\sc patr-ii}),
where any feature can be added to any structure, ProFIT follows the notion
of appropriateness in Carpenter's logic of typed feature structures
\cite{Carpenter:92}, and
introduces features for particular sorts. For each sort, one must declare
which features are introduced by it. The features introduced by a
sort are inherited by all its subsorts, which may also introduce additional
features. A feature must be introduced only at one most general sort.
This makes it possible to provide a notation in which the sort name can
be omitted since it can be inferred from the use of a feature that is
appropriate for that sort.

This notion of appropriateness
is desirable for structuring linguistic knowledge, as it prevents the
ad-hoc introduction of features, and requires a careful design of the
sort and feature hierarchy. Appropriateness is also a prerequisite for
compilation of feature terms into fixed-arity Prolog terms.

Each feature
has a sortal restriction for its value. If a feature's value is only restricted
to
be of sort {\tt top}, then the sortal restriction can be omitted.
The syntax of feature declarations is given in (\ref{feat-decl}).

\begin{equation}
\begin{array}{lll}
Sort & {\tt \ intro\ } & [Feature_1:Restr_1, \\
     &                 & \hspace*{14mm} \vdots \\
     &                 & Feature_n:Restr_n].
\end{array}
\label{feat-decl}
\end{equation}

The following declaration defines a sort {\em binary\_tree\/}
with subsorts {\em leaf\/} and {\em internal\_node\/}. The sort
{\em binary tree\/} introduces the feature {\em label\/} and
its subsort adds the features {\em left\_daughter\/} and
{\em right\_daughter\/}.
If a sort has subsorts and introduces features, these are combined in
one declaration.

\begin{verbatim}
binary_tree >  [leaf,internal_node]
   intro [label].

internal_node
     intro [left_daughter:binary_tree,
            right_daughter:binary_tree].
\end{verbatim}

\subsection{Sorted Feature Terms}

On the basis of the declarations, sorted feature terms can be used in definite
clauses in addition to and in combination with Prolog terms. A Prolog term
can have a feature term as its argument, and a feature can have a Prolog
term as its value. This avoids potential interface problems
between different representations, since terms do not have to be translated
between different languages. As an example, semantic representations in
first-order terms can be used as feature values, but do not need to be
encoded as feature terms.

Sorted feature terms consist of a specification of the sort of the term
(\ref{sort}), or the specification of a feature value (\ref{fv}), or
a conjunction of terms (\ref{conj}). A complete BNF of all ProFIT terms is
given in the appendix.

\begin{eqnarray}
< \mbox{\em Sort}         \label{sort}  \\
\mbox{{\em Feature\/} ! \em Value}  \label{fv}    \\
\mbox{{\em Term\/} \& \em Term}   \label{conj}
\end{eqnarray}

The following clauses (based on {\sc hpsg}) state that a structure is
saturated if its subcat value is the empty list, and that a structure
satisfies the Head Feature Principle ({\tt hfp}) if its head features are
identical with the head features of its head daughter.%
\footnote{These clauses assume appropriate declarations for the
          sort {\tt elist}, and for the
          features {\tt synsem, local, cat, subcat, head, dtrs} and
          {\tt head\_dtr.}}
Note that these clauses provide a concise notation because uninstantiated
features can be omitted, and the sorts of structures do not have to be
specified explicitly because they can be infered from use of the features.

{\small
\begin{verbatim}
saturated( synsem!local!cat!subcat!<elist ).

hfp( synsem!local!cat!head!X &
     dtrs!head_dtr!synsem!local!cat!head!X ).
\end{verbatim}
}

Note that conjunction also provides the possiblity to tag a Prolog term
or feature term with a variable ({\tt Var} \& {\tt Term}).

\subsection{Feature Search}

In the organisation of linguistic knowledge, feature structures are
often deeply embedded, due to the need to group together sets of features
whose value can be structure-shared. In the course of grammar development,
it is often necessary to change the ``location'' of a feature in order
to get the right structuring of information.

Such a  change of the ``feature geometry'' makes it necessary to change the
path in all references to a feature. This is often done by introducing
templates whose sole purpose is the abbreviation of a path to a feature.

ProFIT provides a mechanism to search for paths to features automatically
provided that the sortal restrictions for the feature values are strong
enough to ensure that
there is a unique minimal path. A path is minimal if it does not
contain any repeated features or sorts.

The sort from which to start the feature search must either be specified
explicitly (\ref{search1}) or implicitly given through the sortal restriction
of a feature value, in which case the sort can be omitted and the expression
(\ref{search2}) can be used.

\begin{eqnarray}
%\mbox{\tt Sort>>>Feature ! Term} \label{search1}  \\
%\mbox{\tt >>>Feature ! Term}     \label{search2}
\mbox{{\em Sort\/} $>>>$ {\em Feature\/} ! {\em Term\/}} \label{search1}  \\
>>> \mbox{{\em Feature\/} ! {\em Term\/}}     \label{search2}
\end{eqnarray}

The following clause makes use of feature search to express the Head
Feature Principle ({\tt hfp}).

\begin{verbatim}
hfp( sign>>>head!X &
     dtrs!head_dtr! >>>head!X ).
\end{verbatim}

While this abbreviation for feature paths is new for formal description
languages, similar abbreviatory conventions are often used in linguistic
publications. They are easily and unambiguously understood if
there is only one unique path to the feature which is not embedded in
another structure of the same sort.

\subsection{Templates}

The purpose of templates is to give names to frequently used structures.
In addition to being an abbreviatory device, the template mechanism
serves three other purposes.

\begin{itemize}
\item Abstraction and interfacing by providing a fixed name for a value
      that may change,
\item Partial evaluation,
\item Functional notation that can make specifications easier to understand.
\end{itemize}

Templates are defined by expressions of the form (\ref{template}), where
{\em Name} and {\em Value} can be arbitrary ProFIT terms, including
variables, and template calls. There can be several template
definitions with the same name on the left-hand side (relational
templates). Since templates are expanded at
compile time, template definitions must not be
recursive.

\begin{equation}
\mbox{{\em Name} := \em Value.}  \label{template}
\end{equation}

Templates are called by using the template name prefixed with
{\tt @} in a ProFIT term.

Abstraction makes it possible to change data structures by
changing their definition only at one point. Abstraction also
ensures that databases (e.g. lexicons) which make use of these abstractions
can be re-used in different kinds of applications where different
datastructures represent these abstractions.

Abstraction through templates is also useful for defining interfaces
between grammars and processing modules. If semantic processing must
access the semantic representations of different grammars, this can be
done if the semantic module makes use of a template defined
for each grammar that indicates where in the feature structure the semantic
information is located, as in the following example for {\sc hpsg}.

{\small
\begin{verbatim}
semantics(synsem!local!cont!Sem) := Sem.
\end{verbatim}
}

Partial evaluation is achieved when a structure (say a principle of a
grammar) is represented by a template that gets expanded at compile time,
and does not have to be called as a goal during processing.

We show the use of templates for providing functional notation by a simple
example, in which the expression {\tt @first(X)} stands for the first
element of list X, and  {\tt @rest(X)} stands for the tail of list X, as
defined by the following template definition.

\begin{verbatim}
first([First|Rest]) := First.
rest([First|Rest])  := Rest.
\end{verbatim}

The member relation can be defined with the following clauses, which
correspond very closely to the natural-language statement of the member
relation given as comments. Note that expansion of the templates yields
the usual definition of the member relation in Prolog.

\small
\begin{verbatim}
% The first element of a list
% is a member of the list.
member(@first(List),List).

% Element is a member of a list
% if it is a member of the rest of the list
member(Element,List) :-
     member(Element,@rest(List)).
\end{verbatim}
\normalsize

The expressive power of an n-place template is the same as
that of an n+1 place fact.

\subsection{Disjunction}

Disjunction in the general case cannot be encoded in a Prolog term
representat\-ion.%
\footnote{see the complexity analysis by Brew \cite{Brew:91}.}
Since a general treatment of disjunction would involve too much computational
overhead, we provide disjunctive terms only as syntactic sugar. Clauses
containing disjunctive terms are compiled to several clauses, one for
each consistent combination of disjuncts. Disjunctive terms make it
possible to state facts that belong together in one clause, as the
following formulation of the Semantics Principle
({\tt sem\_p}) of {\sc hpsg}, which states
that the content value of a head-adjunct structure is the content value
of the adjunct daughter, and the content value of the other headed
structures (head-complement, head-marker, and head-filler structure)
is the content value of the head daughter.

\small
\begin{verbatim}
sem_p(  (<head_adj &
         >>>cont!X & >>>adj_dtr!>>>cont!X )
      or
        ( (   <head_comp
           or <head_marker
           or <head_filler
          ) &
          >>>cont!Y & >>>head_dtr!>>>cont!Y )
     ).
\end{verbatim}
\normalsize

For disjunctions of atoms, there exists a Prolog term representation,
which is described below.

\subsection{Finite Domains} \label{sec:findom}

For domains involving only a finite set of atoms as possible values,
it is possible to provide a Prolog term representation (due to Colmerauer,
and described by Mellish \cite{Mellish:88}) to encode any
subset of the possible values in one term.

Consider the agreement features {\em person\/} (with values {\tt 1, 2} and
{\tt 3}) and {\em number\/}
(with values {\tt sg} and {\tt pl}). For the two features together there
are six possible combinations of values (1\&sg, 2\&sg, 3\&sg, 1\&pl, 2\&pl,
3\&pl). Any subset of this set of possible values can be encoded as one
Prolog term. The following example shows the declaration needed for this
finite domain, and some clauses that refer to subsets of the possible
agreement values by making use of the
logical connectives {\tt \~\  }(negation), \& (conjunction),
{\tt or} (disjunction).%
\footnote{The~syntax for finite domain terms is {\tt Term@Domain}.
          However, when atoms from a finite domains are combined by the
          conjunction, disjunction and negation connectives, the
          specification of the domain can be omitted. In the example,
          the domain must only be specified for the value {\tt 2},
          which could otherwise be confused with the integer 2.}

{\small
\begin{verbatim}
        agr fin_dom [1,2,3] * [sg,pl].

        verb(sleeps,3&sg).
        verb(sleep, ~(3&sg)).
        verb(am,    1&sg).
        verb(is,    3&sg).
        verb(are,   2 or pl).

        np('I',     1&sg).
        np(you,     2@agr).
\end{verbatim}
}

This kind of encoding is only applicable to domains which have no
coreferences reaching into them, in the example only the agreement
features as a whole
can be coreferent with other agreement features, but not the values of
person or number in isolation. This kind of encoding is useful to
avoid the creation of choice points for
the lexicon of languages where one inflectional form may correspond
to different feature values.

\subsection{Cyclic Terms}

Unlike Prolog, the concrete syntax of ProFIT allows to write down
cyclic terms by making use of conjunction:

{\small
\begin{verbatim}
        X & f(X).
\end{verbatim}
}

Cyclic terms constitute no longer a theoretical or practical
problem in logic programming, and almost all modern Prolog implementations
can perform their unification (although they can't print them out).
Cyclic terms arise naturally
in NLP through unification of non-cyclic terms, e.g., the Subcategorization
Principle and the Spec Principle of {\sc hpsg}.

ProFIT supports cyclic terms by being able to print them out as
solutions. In order to do this, the dreaded occur check must be performed.
Since this must be done only when results are printed out
as ProFIT terms, it does not affect the runtime performance.

\section{From ProFIT terms to Prolog terms}

\subsection{Compilation of Sorted Feature Terms}

The compilation of sorted feature terms into a Prolog term representation
is based on the following principles, which are explained in more detail
in \cite{Mellish:88,Mellish:92,Schoeter:93,Erbach:94b}.

\begin{itemize}
\item The Prolog representation of a sort is an instance of the
      Prolog representation of its supersorts.
\item Features are represented by arguments. If a feature is introduced
      by a subsort, then the argument is added to the term that further
      instantiates its supersort.
\item Mutually exclusive sorts have different functors at the same
      argument position, so that their unification fails.
\end{itemize}

We illustrate these principles for compiling sorted feature terms
into Prolog terms with an example from {\sc hpsg}. The following declaration
states that the sort {\tt sign} has two mutually exclusive
subsorts {\tt lexical} and {\tt phrasal} and introduces four features.

\small
\begin{verbatim}
sign > [lexical,phrasal]
     intro [phon,
            synsem,
            qstore,
            retrieved].
\end{verbatim}
\normalsize

In the corresponding Prolog term representation below, the first argument is
a variable whose only purpose is being able to test whether two terms
are coreferent or whether they just happen to have the same sort and
the same values for all features. In case of extensional sorts (see
section \ref{sort-declarations}), this variable is omitted. The second
argument can be further instantiated for the subsorts, and the remaining
four arguments correspond to the four features.

\small
\begin{verbatim}
$sign(Var,LexPhras,Phon,Synsem,Qstore,Retriev)
\end{verbatim}
\normalsize

The following declaration introduces two sort hierarchy ``dimensions''
for subsorts of {\tt phrasal}, and one new feature. The corresponding Prolog
term
representation instantiates the representation for the sort {\tt sign}
further, and leaves argument positions that can
be instantiated further by the subsorts of {\tt phrasal}, and for the newly
introduced feature {\tt daughters}.

\small
\begin{verbatim}
phrasal > [headed,non_headed] * [decl,int,rel]
     intro [daughters].
\end{verbatim}

\footnotesize{
\begin{verbatim}
$sign(Var,
      $phrasal(Phrasesort,Clausesort,Dtrs),
      Phon,
      Synsem,
      Qstore,
      Retrieved)
\end{verbatim}
}
\normalsize

\subsection{Compilation of Finite Domains}

The compilation of finite domains into Prolog terms is performed by the
``brute-force'' method described
in \cite{Mellish:88}. A finite domain with $n$ possible domain elements
is represented by a Prolog term with $n+1$ arguments. Each domain
element is associated with a pair of adjacent arguments. For example,
the agreement domain {\tt agr} from section~\ref{sec:findom} with its
six elements (1\&sg, 2\&sg, 3\&sg, 1\&pl, 2\&pl, 3\&pl) is represented
by a Prolog term with seven arguments.

\begin{verbatim}
$agr(1,A,B,C,D,E,0)
\end{verbatim}

Note that the first and last argument {\em must} be different. In the
example, this is achieved by instantiation with different atoms, but an
inequality constraint (Prolog {\sc ii}'s {\tt dif}) would serve the
same purpose. We assume that the domain element 1\&sg corresponds
to the first and second arguments, 2\&sg to the second and third arguemnts,
and so on, as illustrated below.

\begin{tabbing}
\tt
\$agr(      1 \= \hspace*{0.5mm} , \hspace*{0.5mm}
        \tt A \= \hspace*{0.5mm} , \hspace*{0.5mm}
        \tt B \= \hspace*{0.5mm} , \hspace*{0.5mm}
        \tt C \= \hspace*{0.5mm} , \hspace*{0.5mm}
        \tt D \= \hspace*{0.5mm} , \hspace*{0.5mm}
        \tt E \= \hspace*{0.5mm} , \hspace*{0.5mm}
        \tt 0  ) \\
\rm
          \> \hspace*{-1mm} 1sg  \> \hspace*{-1mm} 2sg  \> \hspace*{-1mm} 3sg
\> \hspace*{-1mm} 1pl  \> \hspace*{-1mm} 2pl  \> \hspace*{-1mm} 3pl
\end{tabbing}

A domain description is translated into a Prolog term by unifying the
argument pairs that are {\em excluded} by the description.
For example, the domain description
{\tt 2~or~pl} excludes 1\&sg and 3\&sg, so that the the first and second
argument are unified (1\&sg), as well as the third and fourth (3\&sg).

\begin{verbatim}
$agr(1,1,X,X,D,E,0)
\end{verbatim}

When two such Prolog terms are unified, the union of their excluded
elements is computed by unificatation, or conversely the intersection
of the elements which are in the domain description. The unification
of two finite domain terms is successful as long as they have at least
one element in common. When two terms are unified which have no element
in common, i.e., they {\em exclude} all domain elements, then unification
fails because all arguments become unified with each other, including
the first and last arguments, which are different.

\section{Implementation}

ProFIT has been implemented in Quintus and Sicstus Prolog, and should
run with any Prolog that conforms to or extends the proposed ISO Prolog
standard.

All facilities needed for the development of application programs, for
example the module system and declarations (dynamic, multifile etc.)
are supported by ProFIT.

Compilation of a ProFIT file generates two kinds of files as output.
\begin{enumerate}
\item Declaration files that contain information for compilation, derived
      from the declarations.
\item A program file (a Prolog program) that contains the clauses, with all
      ProFIT terms compiled into their Prolog term representation.
\end{enumerate}

The program file is compiled on the basis of the declaration files. If the
input and output of the program (the exported predicates of a module) only
make use of Prolog terms, and feature terms are only used for internal
purposes, then the program file is all that is needed. This is for example
the case with a grammar that uses feature terms for grammatical description,
but whose input and output (e.g. graphemic form and logical form) are
represented as normal Prolog terms.

Declarations and clauses can come in any order in a ProFIT file, so that
the declarations can be written next to the clauses that make use of them.
Declarations, templates and clauses can be distributed across several
files, so that
it becomes possible to modify clauses without having to recompile the
declarations, or to make changes to parts of the sort hierarchy without
having to recompile the entire hierarchy.

Sort checking can be turned off for debugging purposes, and feature
search and handling of cyclic terms can be turned off
in order to speed up the compilation process if they are not needed.

Error handling is currently being improved to give informative and helpful
warnings in case of undefined sorts, features and templates, or cyclic
sort hierarchies or template definitions.

For the development of ProFIT programs and grammars, it is necessary
to give input and output and debugging information in ProFIT terms,
since the Prolog term representation is not very readable. ProFIT
provides a user interface which

\begin{itemize}
\item accepts queries containing ProFIT terms, and translates them into
      Prolog queries,
\item converts the solutions to the Prolog query back into ProFIT terms
      before printing them out,
\item prints out debugging information as ProFIT terms.
\end{itemize}

When a solution or debugging information is printed out, uninstantiated
features are omitted, and shared structures are printed only once and
represented by variables on subsequent occurences.

A pretty-printer is provided that produces a neatly formatted screen
output of ProFIT terms, and is configurable by the user. ProFIT terms
can also be output in \LaTeX\ format, and an interface to the graphical feature
editor Fegramed is foreseen.

In order to give a rough idea of the efficiency gains of a compilation
into Prolog terms instead of using a feature term unification algorithm
implemented on top of Prolog, we have compared the runtimes
with ALE and the Eisele-D\"orre algorithm for unsorted
feature unification for the following tasks:
(i) unification of (unsorted) feature structures,
(ii) unification of inconsistent feature structures (unification failure),
(iii) unification of sorts,
(iv) lookup of one of 10000 feature structures (e.g. lexical items),
(v) parsing with an {\sc hpsg} grammar to provide a mix of the above tasks.

The timings obtained so far indicate that ProFIT is 5 to 10 times faster
than a system which implements a unification algorithm on top of Prolog,
a result which is predicted by the studies of Sch\"oter \cite{Schoeter:93}
and the experience of the Core Language Engine.

%\begin{table}[h]
%\begin{tabular}{lrrr}
%Problem                           & ProFIT & ALE  & Eisele-D\"orre algorithm
%%\\
%Unification of feature structures &      1 &      & 3-5
%%\\
%Unification failure               &      1 &      & 5
%%\\
%Unification of sorts              &      1 &      & ---
%%\\
%Lookup of lexical entry           &      1 &      & 13
%%\\
%{\sc hpsg} parsing                      &      1 & 5-10 &
%\end{tabular}
%\end{table}

The ProFIT system and documentation are available free of charge by anonymous
ftp
(server: ftp.coli.uni-sb.de, directory: pub/profit).

\section{Conclusion}

ProFIT allows the use of sorted feature terms in Prolog programs and Logic
Grammars without sacrificing the efficiency of Prolog's term unification.
It is very likely that the most efficient commercial
Prolog systems, which provide a basis for the implementation of NLP
systems, will conform to the proposed ISO standard. Since the ISO
standard includes neither inheritance hierarchies nor feature terms
(which are indispensible for the development of large grammars, lexicons
and knowledge
bases for NLP systems), a tool like ProFIT that compiles sorted feature
terms into Prolog terms is useful for the development of grammars and
lexicons that can be used for applications. ProFIT is not a grammar
formalism, but rather aims to extend current and future formalisms
and processing models in the logic grammar tradition with the
expressive power of sorted feature terms. Since the output of ProFIT
compilation are Prolog programs, all the techniques developed for
the optimisation of logic programs (partial evaluation, tabulation,
indexing, program transformation techniques etc.) can be applied
straightforwardly to improve the performance of sorted feature grammars.

\section{Acknowledgements}
This work was supported by
\begin{itemize}
\item The Commission of the European Communities through the project
LRE-61-061 ``Reusable Grammatical Resources'', where it has been (ab-)used in
creative ways to prototype extensions for the ALEP formalism such as
set descriptions, linear precedence constraints and guarded
constraints \cite{Manandhar:94,Manandhar:95}.

\item Deutsche Forschungsgemeinschaft,
Special Research Division 314 ``Artificial Intelligence - Knowledge-Based
Systems'' through project N3 ``Bidirectional Linguistic Deduction'' (BiLD),
where it is used to compile typed feature grammars into logic grammars,
for which bidirectional NLP algorithms are developed, and

\item Cray Systems (formerly PE-Luxembourg), with whom we had fruitful
interaction concerning the future development of the ALEP system.
\end{itemize}

Some code for handling of finite domains was adapted from a program by
Gertjan van Noord. Wojciech Skut and Christian Braun were a great help
in testing and improving the system. Thanks to all the early users and
$\beta$-testers for discovering bugs and inconsistencies, and for
providing feedback and encouragement. Special thanks for {\small \tt service
with a smiley :-)}.

%\bibliography{gorsrev}
%\bibliographystyle{acl}

\vspace*{15mm}
\hspace*{-83mm}
\begin{minipage}{160mm}
\noindent
\rule{153mm}{0.1mm}
\vspace*{10mm}
\appendix
\section*{Appendix: BNF for ProFIT Terms}
\vspace*{10mm}
{\small
\begin{verbatim}
PFT :=  <Sort                 [1.  Term of a sort Sort                        ]
      | Feature!PFT           [2.  Feature-Value pair                         ]
      | PFT & PFT             [3.  Conjunction of terms                       ]
      | PROLOGTERM            [4.  Any Prolog term                            ]
      | FINDOM                [5.  Finite Domain term, BNF see below          ]
      | @Template             [6.  Template call                              ]
      | ` PFT                 [7.  Quoted term, is not translated             ]
      | `` PFT                [8.  Double-quoted, main functor not translated ]
      | >>>Feature!PFT        [9.  Search for a feature                       ]
      | Sort>>>Feature!PFT    [10. short for <Sort & >>>Feature!PFT           ]
      | PFT or PFT            [11. Disjunction; expands to multiple terms     ]
\end{verbatim}
}

{\small
\begin{verbatim}

FINDOM :=  FINDOM@FiniteDomainName
         | ~FINDOM
         | FINDOM & FINDOM
         | FINDOM or FINDOM
         | Atom
\end{verbatim}
}
\end{minipage}
\end{document}